\documentclass[runningheads]{llncs}
\usepackage[T1]{fontenc}
\usepackage{graphicx}
\usepackage{amsmath,amssymb,amsfonts}
\usepackage{algorithmic}
\usepackage{graphicx}
\usepackage{textcomp}
\usepackage{xcolor}

\usepackage{times}
\usepackage{soul}
\usepackage{url}
\usepackage[hidelinks]{hyperref}
\usepackage[utf8]{inputenc}
\usepackage[small]{caption}
\usepackage{booktabs}
\usepackage{algorithm}
\usepackage{algorithmic}
\usepackage[switch]{lineno}

\usepackage[normalem]{ulem}
\usepackage{enumitem}

\newlist{RQ}{enumerate}{1}
\setlist[RQ,1]{label=\textbf{RQ\arabic*:}}

\useunder{\uline}{\ul}{}

\begin{document}
\title{Evaluating LLMs Capabilities Towards Understanding Social Dynamics}
%
%
\author{Anique Tahir\inst{1}\orcidID{0000-0002-2838-147X} \and
Lu Cheng\inst{2}\orcidID{0000-0002-2503-2522} \and \\
Manuel Sandoval\inst{3}\orcidID{0009-0009-8590-8811}  \and
Yasin N. Silva\inst{3}\orcidID{0000-0003-1852-1683}  \and \\
Deborah L. Hall\inst{4}\orcidID{0000-0003-2450-3596}  \and
Huan Liu\inst{1}\orcidID{0000-0002-3264-7904}}
\authorrunning{A. Tahir et al.}
%
\institute{School of Computing and AI, Arizona State University, Tempe AZ\\
\email{\{artahir, huanliu\}@asu.edu}
\and
University of Illinois Chicago, Chicago IL \\
\email{lucheng@uic.edu} \and
Loyola University Chicago, Chicago IL\\
\email{\{msandovalmadrigal, ysilva1\}@luc.edu} \and
School of Social \& Behavioral Sciences, Arizona State University, Glendale AZ \\
\email{d.hall@asu.edu}
}
\maketitle              
\begin{abstract}
Social media discourse involves people from different backgrounds, beliefs, and motives. Thus, often such discourse can devolve into toxic interactions. Generative Models, such as Llama and ChatGPT, have recently exploded in popularity due to their capabilities in zero-shot question-answering. Because these models are increasingly being used to ask questions of social significance, a crucial research question is whether they can understand social media dynamics. This work provides a critical analysis regarding generative LLM’s ability to understand language and dynamics in social contexts, particularly considering cyberbullying and anti-cyberbullying (posts aimed at reducing cyberbullying) interactions. Specifically, we compare and contrast the capabilities of different large language models (LLMs) to understand three key aspects of social dynamics: language, directionality, and the occurrence of bullying/anti-bullying messages. We found that while fine-tuned LLMs exhibit promising results in some social media understanding tasks (understanding directionality), they presented mixed results in others (proper paraphrasing and bullying/anti-bullying detection). We also found that fine-tuning and prompt engineering mechanisms can have positive effects in some tasks. We believe that a understanding of LLM’s capabilities is crucial to design future models that can be effectively used in social applications.

\end{abstract}

\section{Introduction}

In today's digitally connected world, social media platforms have become integral arenas for communication, allowing individuals from diverse backgrounds to engage in discussions, share ideas, and express their opinions. However, this increased connectivity has brought with it a range of challenges, including the emergence of toxic online behavior, cyberbullying, and the propagation of harmful content~\cite{atske_state_2021}. As the online landscape becomes more complex and dynamic, the need to understand and address these issues at scale becomes ever more pressing.

While previous research has made significant strides in specific social interaction settings such as identifying and combating online toxicity and bullying ~\cite{lan_effectiveness_2022}, a notable gap remains in understanding the positive forces at play within digital conversations. Generative models such as Llama and ChatGPT are increasingly being viewed as a panacea for arbitrary problems. However, employing them in social settings can have detrimental consequences. Thus, this paper aims to investigate the extent to which language models can classify and provide explanations for their decisions when social factors are involved. Since the area of social dynamics is diverse, we use bullying and anti-bullying as exemplars for our analysis. We define anti-bullying as interactions where online mediators intervene in toxic discussions with the goal of counteracting cyberbullying behaviors. In essence, these mediators are engaging in bystander intervention~\cite{doi:10.1177/1461444820902541}. Our analysis seeks to pave the way toward providing interpretable insights into online discourse patterns from the perspective of generative language models, further contributing to the comprehension of human interaction in the digital realm.

Recently, large language models (LLMs) have gained traction for their general problem-solving capabilities. LLMs, such as ChatGPT, have been claimed to show signs of intelligence~\cite{bubeck_sparks_2023} and outperform human crowdsourcing~\cite{he_annollm_2023}. In contrast, several efforts have studied the limitations of LLMs, such as their tendency to hallucinate\cite{ye_cognitive_2023}, regurgitate obsolete information which is attributed to the staleness of their training data, and to have racial, gender, and religious biases~\cite{tamkin_understanding_2021}.



In the context of social dynamics, LLMs can potentially provide a pathway to identify and explain behaviors on a large scale. In addition, they could also explain the rationale behind social discourse and, in turn, enable the promotion of positive interactions. This work aims to identify current strengths and limitations of LLMs to identify and explain social behaviors. Our evaluation framework considers multiple enhancement techniques and aims to answer the following research questions:
\begin{RQ}
    \item Do LLMs understand language in the social context?
    \item Which evaluation dimensions expose the weaknesses of LLMs in a social analysis setting?
    \item Can LLMs understand directionality in the social context?
    \item Can LLMs identify behaviors involved in social dynamics, such as cyberbullying and anti-bullying?
\end{RQ}

\section{Related Work}
There is considerable prior research that studies social dynamics~\cite{ROSA2019333}. While this paper aims to evaluate general LLMs capabilities in social contexts, we use cyberbullying and anti-bullying as case studies in our analysis. 

\subsection{LLMs in social media analysis}
LLMs have been extensively used in prior work related to social media analysis. Transformer based models~\cite{vaswani2017attention}, such as BERT~\cite{devlin2018bert}, encode words into tokens and learn attention weights signifying relationship between sequences. Although these models were originally developed for machine translation, they have been repurposed for classification tasks.

\subsubsection{Sentiment Analysis and Emotion Detection}
One prominent application of LLMs in social media analysis is sentiment analysis, which involves determining the emotional tone behind a piece of text. Prior researchers have utilized models like BERT and GPT-2 to gauge sentiment polarity in tweets~\cite{mohammad2019semeval,sun2019finbert}. Emotion detection has also been explored, with models being fine-tuned to recognize emotions such as joy, anger, and sadness~\cite{felbo2017using,gaind2019emotion}.

\subsubsection{Irony and Sarcasm Detection}
Identifying irony and sarcasm presents a unique challenge in online discourse due to the absence of vocal cues and facial expressions. LLMs, including RoBERTa, have been leveraged to detect instances of irony and sarcasm in tweets~\cite{nguyen2020bertweet,zotova2021semi}. These models employ contextual understanding to differentiate between literal and figurative language, contributing to a deeper comprehension of social media interactions.

\subsubsection{Hate Speech and Offensive Language Detection}
The issue of hate speech and offensive language on social media platforms has spurred efforts to develop automated tools for detection and mitigation. Researchers have employed LLMs to create models capable of identifying hate speech~\cite{basile2019semeval,malmasi2018challenges} and offensive content~\cite{mandl2020germeval,risch2021overview,davidson2019racial}. The contextual awareness of LLMs aids in distinguishing between genuine expressions of opinion and harmful speech.

\subsection{Improving Generative Language models: Fine-tuning vs. Prompt Engineering}
For a few years now, standard practice has been to pretrain a model using abundantly available textual data, and then fine-tune the model to perform a more specific task using a new task-specific objective function and/or task-specific dataset. This approach gives the model an opportunity to learn the various general-purpose features of the pretraining data, and avoid biases that could result from being trained on a small dataset.  In practice, the task-specific dataset can be difficult to collect, in short supply, or expensive to produce. Prompt engineering addresses the need for these datasets by having the model instead learn from prompts, a text template engineered specifically to get a better response from the language model. The drawback to prompt engineering is the complexity of identifying how to correctly structure the templates for each task the model will perform ~\cite{liu2021pretrain}. 

The expanding repertoire of LLM applications in social media analysis underscores their potential in unraveling the nuanced behaviors on these platforms. We aim to shed light on LLMs’ strengths and areas in need of additional work to effectively understand intricate social dynamics.

\section{Prompt Generation for Detecting and Explaining User behavior}

\subsection{Background}
Many conversations in social media revolve around certain topics of interest. For instance, information on Reddit is structured in a hierarchical format where sub-reddits are centered around a large scale topic and each thread in a subreddit is a subset of the larger topic. The comments in a thread are centered around the thread topic. Similarly, X~(formerly Twitter) uses hashtags to convey several topics of interest. Posters on X can reply to a comment, creating a hierarchical discourse structure. Instagram and 4chan use similar discourse structures. We refer to the combination of a topic and its responses as a session. Prior work in social media analysis has extensively studied the extraction of useful information from social media sessions, ranging from fake news detection~\cite{liu_early_2018,10.1145/3447548.3467321,duzen_misinformation_2022} to echo chamber formation~\cite{10.1145/3178876.3186130,10.1145/3341161.3343689,10.1145/3511808.3557253}.

Given our interest in finding the capabilities of LLMs to detect and explain cyberbullying and anti-bullying behavior, we formalize next our problem setting. Given a dataset of social media discourse containing text-based conversations, posts, comments, or interactions from various social media platforms, a goal of our work is to analyze whether it is feasible to develop an LLM-based framework for identifying behavior within these online discussions. We define cyberbullies as individuals who demonstrate intentionally hostile or toxic behavior directed toward other users. In contrast, we define anti-bullies as individuals who, in response to cyberbullying, actively seek to counter or combat the harmful content, e.g., by steering conversations in a positive direction or fostering constructive interactions. The problem can be formally stated as follows:

\subsubsection{Input}
The input dataset $\mathcal{D} = \{S_1, S_2, \cdots, S_N\}$ consists of $N$ instances of social media sessions, where each instance $d_{ij} \in \mathcal{S}_i$ is as a sequence of text tokens. Each instance $d_i$ includes the $j$'th comment of session $S_i$ and the previous $1$ to $j-1$ comments.  This representation allows the model to rely on the context of a session.

\subsubsection{Output}
The objective is to produce the following output components:

\begin{enumerate}
    \item \textbf{Behavior Identification (Quantitative):} For each instance $d_{ij}$, let $b_{ij} \in \{0, 1\}$ represent a binary classification of the behavior as a cyberbully. Similarly, we let $m_{ij} \in \{0, 1\}$ represent a binary classification of behavior as anti-bullying.
    
    \item \textbf{Behavior and Motivation Explanations (Qualitative):} For each identified cyberbullying and anti-bullying comment, $b_{ij}, m_{ij} \in \mathcal{S}_i$ respectively, generate human-interpretable explanations that shed light on the behavior, actions, and motivations behind the comment.
\end{enumerate}


Generative LLMs have demonstrated general purpose capabilities in question answering, surpassing humans in some cases~\cite{guerra_gpt-4_2023}. Several methods focused on enhancing the output of these models have been proposed. For our task of social behavior analysis, we created a taxonomy of methods for enhancing responses. The most common approach for querying foundational LLMs is to use zero-shot prompting. However, because these models are trained on general data, they may fail to perform niche tasks. In addition, because natural language is ambiguous, the models may respond in a manner that is not expected for a specific use-case. For instance, a model trained as a `comedy bot' might respond in a humorous style, whereas such a style would not be relevant across a broader range of applications.

There are several approaches to guide prompting for enhancing generation. The most common ones are chain-of-thought~(CoT) prompting~\cite{wei2022chain} and exemplar-based prompting~\cite{min2021metaicl}. In chain-of-thought prompting, the prompt is constructed such that the response consists of the rationale for the decision before responding with the final decision in the generation. Exemplar-based generation provides the model with example generations, in essence guiding the model to continue generations in similar fashion.

\subsection{Prompt Generation}
\begin{figure}
    \centering
    \includegraphics[width=0.8\columnwidth]{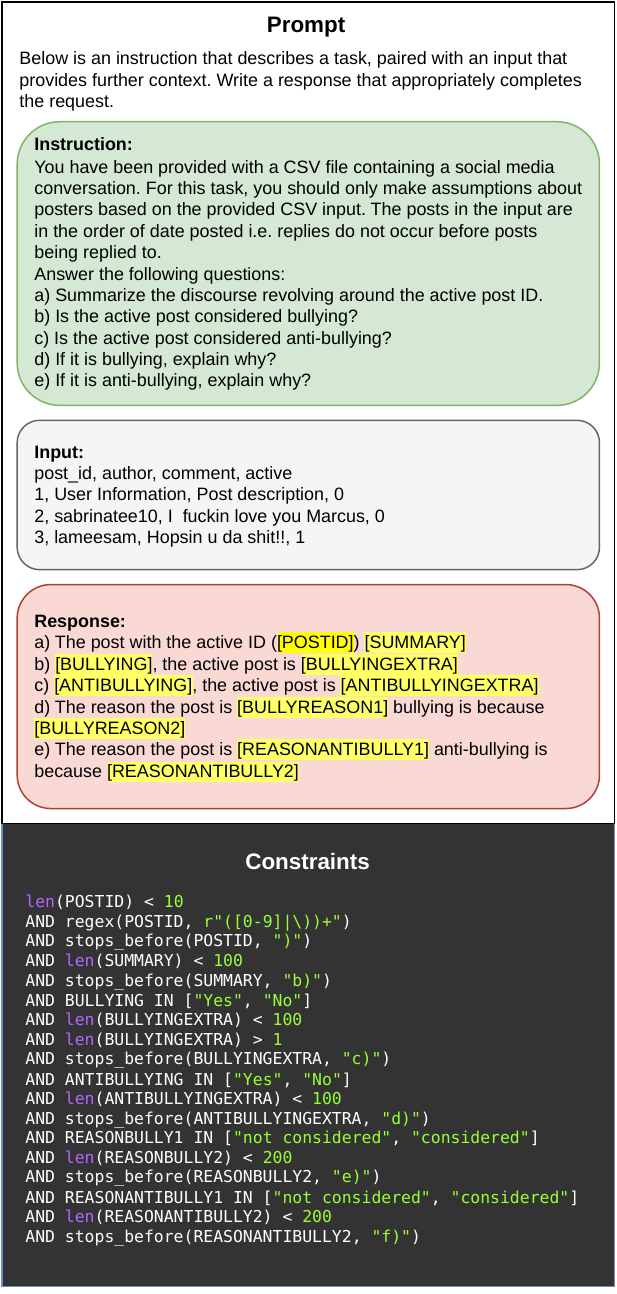}
    \caption{We use constrained generation for our analysis. Constrained prompt generation allows extraction of categorical labels and explanations while preventing the responses from deviating from expectation. The prompt placeholders are highlighted in yellow.}
    \label{fig:prompting}
\end{figure}
\raggedbottom

Because social media sessions have a specific structure, prompt design needs to take four major factors into consideration: (i) The task objective, i.e., identifying and explaining behavior; (ii) context, i.e., what is the topic of the subset of the conversation we are analysing; (iii) the classification instance $d_{ij}$, i.e., how can we differentiate the query objective from the context; and (iv) the response format, i.e., how do we want the LLM to structure its generated response.

Given that a social media conversation is sequential in nature, we use the tabular format such that various characteristics of the conversation are encoded alongside the comments, e.g., the post id, comment author, and a label specifying which comment is the target of classification. In addition, the instruct-tuning prompt format~\cite{touvron_llama_2023} allows the instruction and input to be structured together in a prompt query. Because we are interested in classification, we expect the LLM to return the output in a structured manner such that class labels, $b_{ij}$ and $m_{ij}$, can be extracted from the generated open-ended response. We employ templating of the procedural generation through constraints in the generation process. Fig.~\ref{fig:prompting} shows one of the sample prompts used in our analysis. The first question, which asks for a summary of the conversation around the post of interest, acts as a CoT mechanism.

\subsection{Fine-tuning}
\label{subsec:fine-tuning}
In addition to prompt and response engineering, we also studied the effect of fine-tuning on the veracity of the responses. For this work, each phase of the fine-tuning process is performed using the \textbf{Lo}w \textbf{R}ank \textbf{A}daptation (LoRA) strategy~\cite{hu2021lora}. This strategy allows us to target particular layers in the model for fine-tuning. Consider a weight matrix, $W_0 \in \mathbb{R}^{m \times n}$, representing the original weights. We use a new set of corresponding weights, $A \in \mathbb{R}^{m \times r}$ and $B \in \mathbb{R}^{r \times n}$, where $r << min(m, n)$. Our models' corresponding weight is now $W_0 + AB$. Thus, backpropogating on $A$ and $B$, we can fine-tune on a significantly smaller number of parameters.

Machine learning models are vulnerable to mode collapse and catastrophic forgetting when training on a certain type of data. That is, training on structural understanding might make the model forget about previously learned knowledge in lieu of learning structure. To prevent such a hazard, we use a combination of task-specific data and general instruction data. For the instruction data, we make use of Alpaca~\cite{touvron_llama_2023} instructions, which are combined with the task data with a certain probability. 
\section{Do LLMs understand language in social context?}
\label{sec:social-understanding}

\begin{figure*}[h]
    \centering
    \includegraphics[width=.95\textwidth]{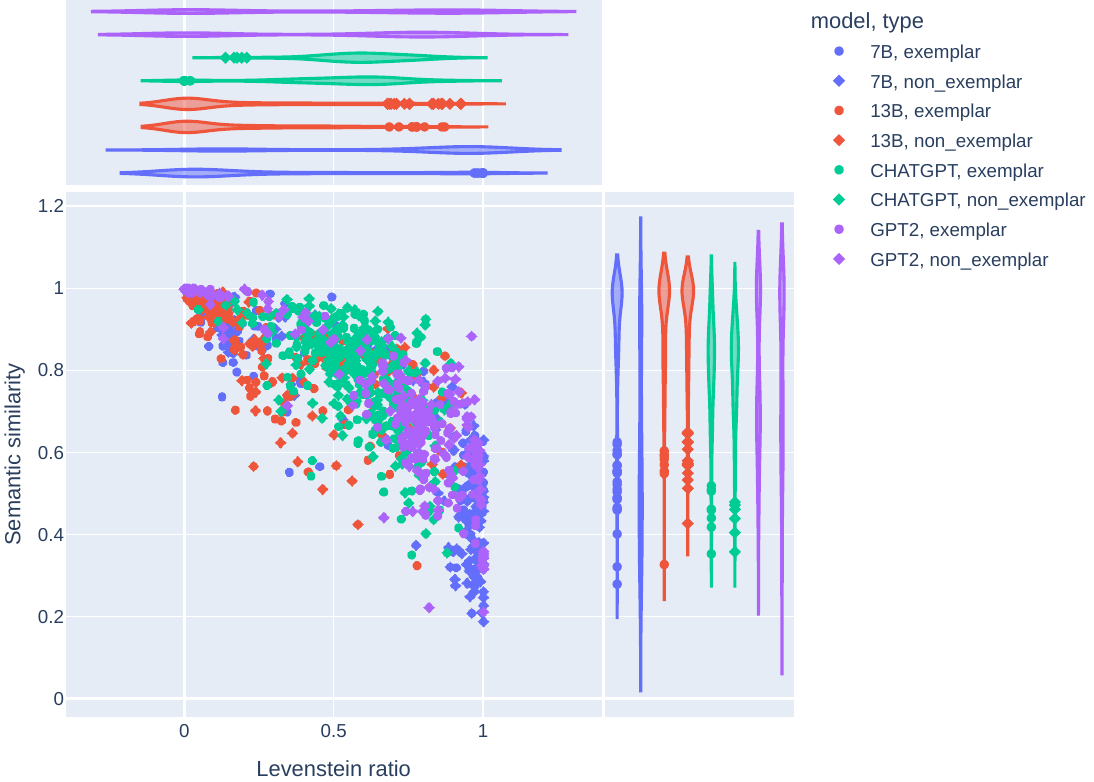}
    \caption{Semantic similarity against edit distance for paraphrases. ChatGPT shows a promising balance, while Llama-based models tend to repeat the text verbatim. Using exemplars shows performance improvements for the Llama-2 7B model.}
    \label{fig:similarity-comparison}
\end{figure*}

Large language models are pretrained mostly on formal and semi-formal text corpora. Given our motivation to analyze social media discourse, we compare the capabilities of LLMs to understand informal social media comments. Our hypothesis is that an LLM that understands a comment will be able to paraphrase it. Comments are used for paraphrasing in two scenarios: (i) instruction-based paraphrasing, and (ii) exemplar-based paraphrasing. Instruction-based paraphrasing consists of an Alpaca format prompt where the instruction asks the language model to paraphrase, the input is the text to paraphrase, and the response is generated by the model. 

\subsection{Similarity Analysis}

We use four main metrics to evaluate the quality of generations: (i) BLEU score, (ii) ROGUE score, (iii) Jaccard similarity, and, (iv) Semantic similarity. One of the quirks with generation is that in many cases, the LLMs reproduce the provided text verbatim. This leads to an increase in most of the similarity metrics. Thus, we also report the Levenshtein ratio for each strata. BLEU score measures the correspondence between the model's output and the reference paraphrase based on n-gram overlap. ROGUE score also evaluates the overlap of n-grams, but with a focus on recall, thus capturing the extent to which the reference n-grams are present in the generated text. Jaccard similarity assesses the similarity and diversity of the sets of n-grams between the generated and reference texts. More specifically, it is the ratio of the intersection and union of the n-grams in the two texts. Semantic similarity explores the likeness in meaning between the texts. In this instance, it is extracted from the encodings of the BERT~(base) model. 

\begin{table*}[]
\centering
\resizebox{\textwidth}{!}{%
\begin{tabular}{@{}lllllllll@{}}
\toprule
               &               & \multicolumn{6}{c}{\textbf{Similarity Metrics}}                                                                                                     & \multicolumn{1}{c}{\textbf{Edit Distance}} \\ \cmidrule(lr){3-8}  \cmidrule(lr){9-9}
\textbf{Model} & \textbf{Type} & \textbf{BLEU}          & \textbf{ROUGE-1}       & \textbf{ROUGE-2}       & \textbf{ROUGE-L}       & \textbf{Jaccard}       & \textbf{Semantic}      & \textbf{Levenshtein}                       \\
13B            & exemplar      & \textbf{0.725 (0.352)} & \textbf{0.865 (0.219)} & \textbf{0.801 (0.299)} & \textbf{0.859 (0.230)} & \textbf{0.784 (0.285)} & \textbf{0.917 (0.124)} & 0.157 (0.228)                              \\
13B            & non\_exemplar  & {\ul 0.687 (0.383)}    & {\ul 0.844 (0.248)}    & {\ul 0.777 (0.322)}    & {\ul 0.838 (0.256)}    & {\ul 0.757 (0.310)}    & {\ul 0.915 (0.128)}    & 0.175 (0.260)                              \\
7B             & exemplar      & 0.654 (0.378)          & 0.768 (0.352)          & 0.734 (0.373)          & 0.763 (0.355)          & 0.678 (0.354)          & 0.888 (0.161)          & 0.239 (0.329)                              \\
7B             & non\_exemplar  & 0.182 (0.310)          & 0.270 (0.396)          & 0.239 (0.393)          & 0.265 (0.396)          & 0.211 (0.313)          & 0.604 (0.233)          & \textbf{0.733 (0.351)}                     \\
CHATGPT        & exemplar      & 0.187 (0.258)          & 0.527 (0.219)          & 0.305 (0.264)          & 0.493 (0.225)          & 0.345 (0.233)          & 0.805 (0.127)          & 0.521 (0.211)                              \\
CHATGPT        & non\_exemplar  & 0.084 (0.094)          & 0.456 (0.172)          & 0.197 (0.183)          & 0.410 (0.165)          & 0.264 (0.137)          & 0.795 (0.119)          & {\ul 0.599 (0.145)}                        \\
GPT2           & exemplar      & 0.401 (0.465)          & 0.489 (0.427)          & 0.426 (0.469)          & 0.479 (0.433)          & 0.451 (0.433)          & 0.788 (0.189)          & 0.488 (0.387)                              \\
GPT2           & non\_exemplar  & 0.453 (0.466)          & 0.532 (0.439)          & 0.480 (0.463)          & 0.526 (0.443)          & 0.522 (0.418)          & 0.800 (0.213)          & 0.467 (0.419)                              \\ \bottomrule
\end{tabular}%
}
\caption{Comparison of similarity metrics and the corresponding edit distance across various language models when paraphrases generated on social media comments are compared with the original comment. Ideal paraphrasing should vary from the original comment but keep the meaning intact.}
\label{tbl:understanding-comparison}
\end{table*}

The previous metrics show the similarity between different text pairs. However, we observed that in many instances, the generations repeated the input verbatim leading to high similarity. Fig.~\ref{fig:similarity-comparison} shows the distribution of similarities in the form of violin plots. To differentiate verbatim or close-to-verbatim generations from generations of higher quality, we also calculate the Levenshtein distance~(edit distance) between the generation and the original text. Because this distance varies depending on the length of the text, in practice, we rely on the Levenshtein ratio:
\begin{equation}
    \text{Levenshtein Ratio}(s_1, s_2) = \frac{\text{Levenshtein Distance}(s_1, s_2)}{max(|s_1|, |s_2|)}
\end{equation}

\subsection{Comparison of Social Understanding among LLMs}
We compare different LLMs to understand their ability to comprehend social conversations. Specifically, GPT-2, Llama-2 7B/13B, and ChatGPT are used in our comprehension analysis. Some of the comments include language of toxic nature that ChatGPT refuses to answer due to ethical constraints. For some of our analysis, we used ChatGPT~(GPT 3.5) generations near the timeframe of its release when the restrictions were comparatively lax. Unfortunately, current versions of ChatGPT do not allow most of these prompts. We compare generations with and without exemplars to gauge any improvement in understanding. For the social media content, we use a dateset of Instagram sessions~\cite{hosseinmardi2015analyzing}. The results are tabulated in Table.~\ref{tbl:understanding-comparison}.

\subsubsection{Common LLM mis-generations}
We identify three common types of generation mistakes made by the models: (i) verbatim generation; (ii) repetition of exemplars; and (iii) gibberish generation. When a model generates responses verbatim, the Levenshein distance is $0$. In some cases, exemplars serve as a double-edged sword, as the model reproduces an example instead of producing novel paraphrasing. In many cases, LLMs produce gibberish unrelated to the query comment or the provided exemplars. In such cases, we observe a low similarity score and a high edit distance since the generated text is semantically dissimilar to the origin text and contains few common sub-sequences.

\subsubsection{Significance of Exemplars}
Adding exemplars to the prompt is a common technique to guide generations toward a specific goal. However, in our analysis, we only see a significant change in generation statistics for the 7B variant of Llama-2. For this case, the Levenshtein distance also decreases, hinting at verbatim repetitions of the input. These changes are evident in the distributions represented in the violin plots in Fig.~\ref{fig:similarity-comparison}.

\subsubsection{Importance of Distributional Analysis}
\label{subsubsection:chatgpt-wins}
Although the semantic scores in Table.~\ref{tbl:understanding-comparison} appear to favor Llama-2 13B, even compared to ChatGPT, qualitative analysis and the distributional stratification in Fig.~\ref{fig:similarity-comparison} better contextualises the tradeoffs between textual and semantic similarity. ChatGPT generations produce text that is consistently semantically similar~($similarity \rightarrow 1$) to the input, but avoids being verbatim repetition~($\text{Levenshtein ratio} \rightarrow 0$) or gibberish~($\text{Levenshtein ratio} \rightarrow 1$). Thus, ChatGPT shows a robust understanding of social context, while GPT-2 tends to produce gibberish. The Llama-2 based models tend to produce a mixture of repetition, valid responses, and gibberish. A possible explanation for this phenomenon could be the pre-training corpus, which includes mostly academic text in the case of Llama-2. The results also suggest that exemplars may be insufficient to improve language model understanding, as they only guide the generation process rather than improving the model's inherent understanding~(encoded in the weights).

For social understanding, we conclude that \textit{even though common metrics such and BLEU and ROGUE allude to language understanding, we find problems with regurgitation and verbatim reproductions. We also find that multi-faceted evaluation paints a more nuanced picture.}
\section{Can language models understand directionality in social contexts?}
Many social platforms enable their users to express whom they target with their comments. A majority of social media platforms provide some variation of the `mention' functionality. In the context of this work, we are specifically interested in the ability of LLMs of follow the directionality pattern. 

\subsection{Language Models and Directionality Detection}
\begin{figure}
    \centering
    \includegraphics[width=\columnwidth]{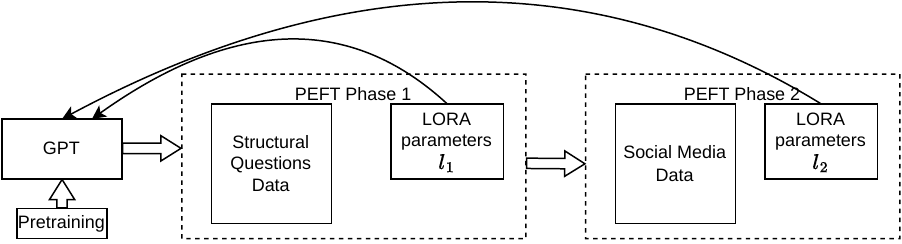}
    \caption{We divide our fine-tuning analysis in two phases. The first phase adds structural understanding, while the second phase adds social understanding to the under-privileged models.}
    \label{fig:method}
\end{figure}

The primary objective of training an LLM for next-word prediction is to learn the conditional probability distribution of the next word given the context of preceding words. Formally, let \(w_1, w_2, \ldots, w_{n-1}\) be a sequence of \(n-1\) words, and \(w_n\) be the target word to predict. The goal is to learn the probability \(P(w_n | w_1, w_2, \ldots, w_{n-1})\), which captures the likelihood of different words occurring next in the sequence. To train LLMs, large text corpora are used as datasets. These corpora contain a wide variety of sentences and documents, providing diverse linguistic contexts for the model to learn from. Common sources of data include books, articles, websites, and other textual sources. 

\subsubsection{LLMs capabilities under fine-tuning}

Because prompt tuning is insufficient for models to develop social understanding, as noted in Section~\ref{sec:social-understanding}, we use a fine-tuning process consisting of two phases to add knowledge to the LLM (illustrated in Fig.~\ref{fig:method}). Both phases of fine-tuning use Parameter Efficient Fine-Tuning~(PEFT)\footnote{See Section \ref{subsec:fine-tuning}}. In our problem setting, rather than just predicting the next words, we aim to gain an understanding of the relation between different comments. For instance, a comment in a session may target the previous comment, the original post that spawned the session, or some comment in the middle of the discourse. To glean insight into the target of the comment in terms of its context, reasoning about the structure of the conversation is critical. Unfortunately, the LLM pre-training does not consider these relationships specifically and there is no public data related to reasoning at the comment level in social media discourse. Thus, we rely on other general purpose structured data to act as a surrogate to learn structure and reasoning. We use the WikiTableQuestions~\cite{pasupat2015compositional} dataset to infuse structural intelligence into the model. This dataset consists of a large variety of independent tables, questions based on one of the tables, and a corresponding answer. To answer a question, it is vital to use the data in the table. 

Phase 2 of the PEFT process aims to improve the social understanding of under-privileged LLMs. ChatGPT shows improved social language understanding compared to competitors\footnote{See Section~\ref{subsubsection:chatgpt-wins}}. Even though, the advantage of ChatGPT may be attributed to a more generalized and customized training corpus, this phase of PEFT seeks to study the transfer learning abilities of competitor models in the confines of our problem setting.

\subsection{Can LLMs emergent abilities improve directional understanding?}

\begin{table}[]
\centering
\resizebox{.65\columnwidth}{!}{%
\begin{tabular}{@{}lccc@{}}
\cmidrule(l){2-4}
                  & \multicolumn{1}{l}{\textbf{Target Post}} & \multicolumn{1}{l}{\textbf{Reply Post}} & \multicolumn{1}{l}{\textbf{p(Reply | Target)}} \\ \cmidrule(l){2-4} 
\textbf{7B}       & 0.082                                    & 0.153                                   & 0.643                                          \\
\textbf{13B}      & 0.159                                    & 0.200                                   & {\ul 0.815}                                    \\
\textbf{7B-PEFT}  & {\ul 0.865}                              & {\ul 0.541}                             & 0.558                                          \\
\textbf{13B-PEFT} & \textbf{0.971}                           & \textbf{0.847}                          & \textbf{0.855}                                 \\ \bottomrule
\end{tabular}%
}
\caption{We find that the veracity of the directionality identification significantly improves with the proposed PEFT fine-tuning phases compared to the base models.}
\label{tab:directionality_analysis}
\end{table}

For the directionality analysis task, we utilized a corpus of 4chan threads~\cite{papasavva2020raiders} and efficient fine-tuning using JORA~\cite{tahir2024jora}. Since 4chan permits its users to tag the individuals to whom they are replying, we employ this data as the ground truth for directionality information. Our objective is to determine whether our designed PEFT phases (i) enhance the model's ability to identify the post being targeted for behavior comprehension, and (ii) improve the model's capacity to discern the individual being targeted by the poster. Given that 4chan users can reference multiple comments as the target of their replies, we consider the model successful in this analysis if it correctly identifies any one of the multiple mentioned comments.

We stratify our analysis based on the number of model parameters and whether the models were fine-tuned. Table~\ref{tab:directionality_analysis} shows the accuracy of directionality identification for the target post and the reply post. The PEFT models significantly outperform the base models. In general, the greater the number of model parameters, the better the veracity of identification. We observe this for the base models as well as the fine-tuned ones. Surprisingly, we find that the base models perform better with reply identification than with target post identification. This behavior is the result of the model relying on randomly mentioning the tags to the other posts in the comments. Because not all posts tag other comments, the probability of returning a valid comment that is tagged at random is greater than the probability of selecting the target comment at random. In the conversation data, the number of comments being replied to is a subset of the total comments. Because PEFT models rely on more educated guesses, the veracity of reply identification is less than the veracity of target identification.

Finally, we examine whether each model can capture the reply direction given that it was able to identify the target. Our results indicate that even the base models yield promising findings for identifying directionality in such a scenario. However, because the probability of identifying the target post is low, this may represent a very small sample of easy examples not representative of the bigger picture. For PEFT models, we see a minor increase in $p(reply_{\text{predicted}} = reply_{\text{ground truth}}|target)$, suggesting that these models have learned the relation between the target and the reply during the fine-tuning phase. In conclusion, \textit{LLMs show promising prospects for learning directionality when they are trained to do so by fine-tuning structural and social intelligence. This result is aligned to recent findings about the emergent abilities of LLMs~\cite{wei2022emergent}.}
\section{Can LLM’s detect instances of Cyberbullying and Anti-Bullying?}
\label{sec:classification}

\begin{table}[]
\centering
\resizebox{0.55\columnwidth}{!}{%
\begin{tabular}{@{}lcccc@{}}
\cmidrule(l){2-5}
                   & \multicolumn{1}{l}{\textbf{7B}} & \multicolumn{1}{l}{\textbf{7B-PEFT}} & \multicolumn{1}{l}{\textbf{13B}} & \multicolumn{1}{l}{\textbf{13B-PEFT}} \\ \cmidrule(l){2-5} 
\textbf{accuracy}  & 0.500                           & 0.481                                & 0.494                            & 0.513                                 \\
\textbf{precision} & 0.500                           & 0.480                                & 0.496                            & 0.514                                 \\
\textbf{recall}    & 0.897                           & 0.462                                & 0.769                            & 0.462                                 \\
\textbf{f1}        & 0.642                           & 0.471                                & 0.603                            & 0.486                                 \\ \bottomrule
\end{tabular}%
}
\caption{Both base and PEFT models show random performance for identification of cyberbullying and anti-bullying comments.}
\label{tbl:classification}
\end{table}


To further validate our findings with respect to understanding social context and to analyze the veracity of behaviour classification as cyberbullying or anti-bullying, we design our experiment on a dataset of Instagram sessions~\cite{hamlett2022labeled} which tests both language and directional understanding. We use 100 labeled sessions. Each session consists of a post and a set of comments. This dataset serves two purposes: (i) Phase 2 of our PEFT is based on 4chan threads. Using a separate dataset allows us to avoid any potential data leakage and measures the generalization capabilities of the model, and (ii) Instagram comments are short compared to 4chan comments. This allows us to add additional context in the input of the prompt. In this experiment, we stratify the evaluation based on the number of model parameters and whether PEFT was applied. 

We partition a random sample of the comments into two sets. The first set consists of comments labeled as cyberbullying and the second set consists of comments labeled as anti-bullying. Comments that fall into neither of the two categories are ignored. This simplifies the analysis to a binary classification problem, which is easier to interpret. Different variations of the LLMs are prompted and allowed to respond in a constrained manner as illustrated in Fig.~\ref{fig:prompting}. Table~\ref{tbl:classification} summarizes the results of our experiment. In terms of accuracy, all variations of the models appear to perform close to random chance. Further investigation of the responses, including the summary, revealed that the models were incapable of understanding the social commentary at a semantic level. Non-PEFT models display a higher recall~(and f1) score since they tend to positively label both cyberbullying and anti-bullying cases. The results are consistent with the findings in Section~\ref{sec:social-understanding} where we show that these models struggle to understand the informal language used in social networks. We were unable to compare with ChatGPT for two reasons. First, ChatGPT is a closed source model and thus does not allow constrained generation. Second, because cyberbullying and anti-bullying comments contain content or context of a toxic nature, safeguards prevent the generation of responses. Given that ChatGPT showed promising ability in informal language understanding, we hypothesize that it would have displayed improved performance in this particular experiment. This highlights the need for, and importance of, a social language comprehension dataset, analogous to SQuAD~\cite{DBLP:journals/corr/abs-1806-03822}, and more robust mechanisms to perform academic evaluations of popular LLMs.

\section{Discussion}
Making inferences on social data is a multi-faceted problem. In this work, we study three dimensions relevant to exploiting LLMs for social behaviour analysis. Foundational models pretrained on large corpora of text, such as Llama, have enabled a larger audience to enjoy their benefits by tuning them for specific use-cases. First, we compare the ability of different models to understand social context by measuring how well they can generate paraphrases. Second, we study whether LLMs can understand directionality in social media. Third, we measure the effectiveness of LLMs in a classification task that uses a combination of directional and social understanding. Our experiments converged in the finding that the weakest link in the considered problem setting is the lack of semantic understanding of the language used in social media. This finding is corroborated by the results in Sections~\ref{sec:social-understanding} and \ref{sec:classification}. In addition, our results show promising ability of LLMs in the domain of directional understanding. Because LLMs are trained on large corpus of semi-formal and formal text, our results show a need for large scale informal training corpuses, preferably focused on language comprehension. 


\section{Conclusion, Limitations, and Future Work}
LLMs have seen a spike in popularity recently due to their astounding ability in zero-shot response generation for general-purpose prompts. There have been various prior studies that either question~\cite{valmeekam2023planning,onoe2023can} or praise~\cite{kiciman2023causal,chen2023can} their abilities in different domains. In this work, we focus on the subdomain of social analysis through the lens of social media discourse. More specifically, we considered the detection of cyberbullying and anti-bullying behavior. We highlight strengths and limitations throughout a taxonomy of learning in language models through prompt engineering, fine-tuning, and constrained generation. We apply relevant parts of this taxonomy to answer research questions related to social context understanding, directionality understanding, and identification of cyberbullying and anti-bullying activity, which requires both social and directional understanding. We believe future breakthroughs in this domain will require larger coded social datasets as well as new model features that better capture the complex nature of social interaction semantics.



\begin{credits}
\subsubsection{\ackname} This work was supported by NSF Awards \#2227488 and \#1719722 and a Google Award for Inclusion Research. Cheng was supported by a Cisco Faculty Research Award.
\end{credits}
%
%
%
\bibliographystyle{splncs04}
\bibliography{main}

\end{document}